\pdfoutput=1

\documentclass[11pt]{article}

\usepackage{EMNLP2022}

\usepackage{times}
\usepackage{latexsym}

\usepackage[T1]{fontenc}

\usepackage[utf8]{inputenc}

\usepackage{microtype}

\usepackage{inconsolata}

\usepackage{amsmath}
\usepackage{hyperref}
\usepackage{url}
\usepackage{mathrsfs}
\usepackage{multirow}
\usepackage{booktabs}
\usepackage{graphicx}
\usepackage{subcaption}
\usepackage{cleveref}
\usepackage{CJK}
\usepackage{CJKspace}
\usepackage{fdsymbol}
\usepackage{enumitem} 
\usepackage[normalem]{ulem}
\useunder{\uline}{\ul}{}
\crefname{equation}{Eq.}{Eq.}
\crefname{section}{Section}{Sections}
\crefname{subsection}{Section}{Sections}
\crefname{subsubsection}{Section}{Sections}
\crefname{figure}{Figure}{Figures}
\crefname{table}{Table}{Tables}
\crefname{subfigure}{Figure}{Figures}
\crefname{algocf}{Algorithm}{Algorithms}
\usepackage{pifont}
\title{CLIP also Understands Text: Prompting CLIP for Phrase Understanding}

\author{
  An Yan\textsuperscript{\dag},
  Jiacheng Li\textsuperscript{\dag},
  Wanrong Zhu\textsuperscript{\P},
  Yujie Lu\textsuperscript{\P},
  \\
  \textbf{William Yang Wang\textsuperscript{\P},
  Julian McAuley\textsuperscript{\dag}
  }
  \\
  \textsuperscript{\dag}UC San Diego, 
  \textsuperscript{\P}UC Santa Barbara 
  \\
    \{ayan, j9li,jmcauley \}@ucsd.edu \\
    \{wanrongzhu,yujielu,william\}@cs.ucsb.edu 
}

\begin{document}
\maketitle

\begin{abstract}
Contrastive Language-Image Pretraining (CLIP) efficiently learns visual concepts by pre-training with natural language supervision. 
CLIP and its visual encoder have been explored on various vision and language tasks and achieve strong zero-shot or transfer learning performance. 
However, the application of its text encoder solely for text understanding has been less explored. 
In this paper, we 
find that the text encoder of CLIP actually demonstrates strong ability for phrase understanding, and can even significantly outperform popular language models such as BERT with a properly designed prompt. 
Extensive experiments validate the effectiveness of our method across different datasets and domains on entity clustering and entity set expansion tasks.
\end{abstract}

\section{Introduction}
Contrastive Language-Image Pretraining (CLIP) \citep{radford2021learning} is a recent model proposed to learn visual concepts from natural language supervision. It consists of a visual encoder and a text encoder, and learns visual representations by aligning images and text through a contrastive loss.
CLIP has demonstrated strong zero-shot open-set image classification capability with 400 million pre-training image-text pairs crawled from the web.

However, despite its success for computer vision and multimodal tasks~\citep{shen2021much}, few studies explore the application of its text encoder on downstream text understanding tasks.
Recently,~\citet{hsu2022xdbert} has empirically found that CLIP performs poorly on natural language understanding tasks directly. One potential reason is that CLIP is not trained with language modeling losses (e.g., masked language modeling, MLM), which proves to be crucial for language understanding~\citep{Devlin2019BERTPO, liu2019roberta}.
But since the visual encoder has benefited from language supervision, one might naturally ask: does the text encoder also benefit from visual supervision?

In this paper, we show that even though CLIP is pre-trained without explicit token-, word- or phrase-level supervision, with a simple and effective prompting method, the CLIP text encoder can be directly used for phrase understanding, and can significantly outperform popular language models trained with masked language modeling (e.g.~BERT) or even phrase-specific learning objectives such as Phrase-BERT~\citep{Wang2021PhraseBERTIP} and UCTopic~\citep{Li2022UCTopicUC} on several phrase-related datasets from different domains. Specifically, we automatically generate instance-level prompts for each phrase by probing the knowledge of a language model. Then, the text encoder of CLIP encodes phrases with corresponding prompts to obtain final representations. We evaluate these representations directly on two phrase understanding tasks without further fine-tuning.

Consequently, CLIP text encoder achieves an average of 6.4\% absolute improvement (70.3\% vs 76.7\% on accuracy) on the entity clustering task (CoNLL2003, BC5CDR, W-NUT2017) and an improvement of 9.8\% (56.9\% vs 66.7\% on mean average precision) on the entity set expansion task (WIKI) compared with the best performing language models.

\begin{figure*}[t!]
     \centering
     \begin{subfigure}[b]{0.43\textwidth}
         \centering
         \includegraphics[width=\textwidth]{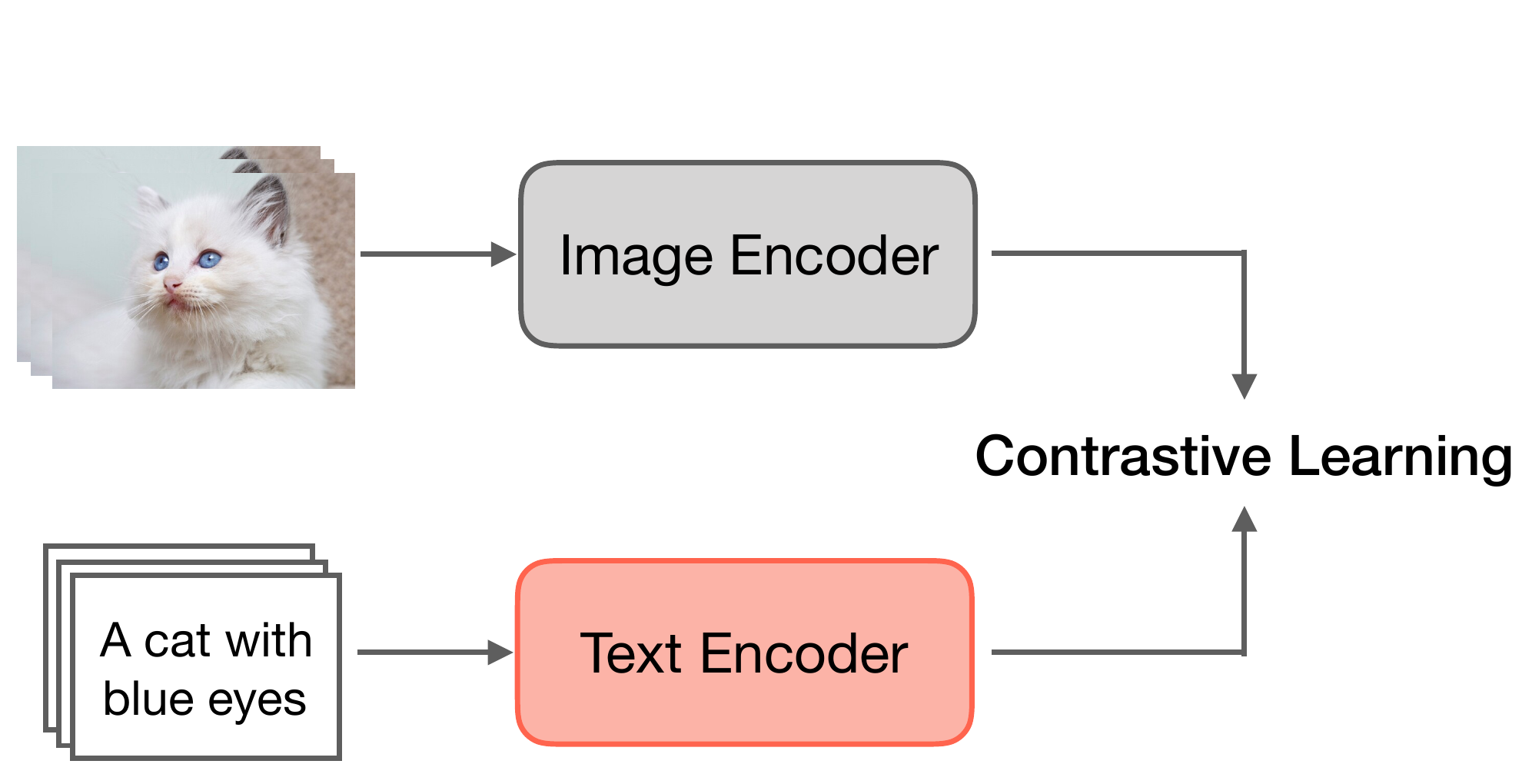}
         \caption{CLIP: Contrastive Language-Image Pretraining}
         \label{fig:clip}
     \end{subfigure}
     ~
     \begin{subfigure}[b]{0.55\textwidth}
         \centering
         \includegraphics[width=\textwidth]{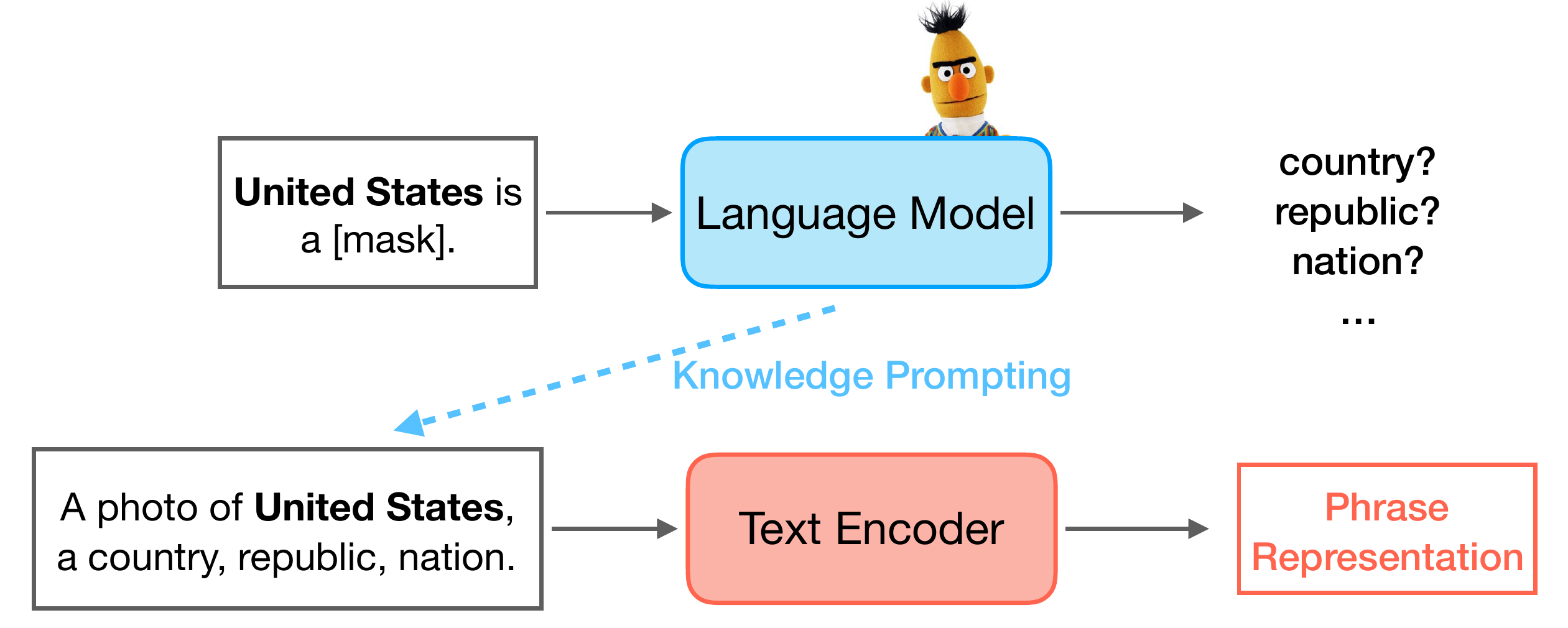}
         \caption{Domain-aware Prompting for CLIP}
         \label{fig:prompt}
     \end{subfigure}
    \caption{Illustration of our framework for phrase understanding with the text encoder of CLIP. 
    }
\end{figure*}

Overall, our contributions are as follows:
\begin{itemize}[noitemsep,topsep=0pt,parsep=0pt,partopsep=0pt]
    \item We are the first to show that a text encoder trained with only image-text contrastive learning  can achieve competitive or even better results on downstream text understanding tasks compared to popular language models pre-trained with MLM.
    \item We design an automatic prompting method with a language model as the knowledge base to boost performance on phrase understanding for both language models and CLIP.
    \item We conduct comprehensive experiments to demonstrate the effectiveness of our method, and analyze why CLIP performs well for these tasks across different domains.
\end{itemize}



\section{Methodology}
\subsection{Preliminary: CLIP}
CLIP is a powerful vision-language model with strong performance for zero-shot open-set image classification. As shown in~\cref{fig:clip}, it consists of two encoders, a ResNET~\citep{he2016deep} or ViT~\citep{dosovitskiy2020image}  image encoder and a transformer text encoder. Given an image and a sequence of words, it will transform them into feature vectors $V$ and $T$ respectively.

Then the model is pretrained with contrastive losses between two modalities:
\begin{equation}
    \mathcal{L}_{\mathit{CLIP}} = \frac{1}{2} (\mathcal{L}_{v\rightarrow t} + \mathcal{L}_{t\rightarrow v})
    \label{eqn:clip}
\end{equation}
where given a mini-batch of $N$ samples, $\mathrm{sim}$ as the cosine similarity, $\tau$ as the temperature, the contrastive loss for $\mathcal{L}_{v\rightarrow t}$ ( similar definitions for $\mathcal{L}_{t\rightarrow v}$) is formulated as:
\begin{equation}
    \mathcal{L}_{v\rightarrow t} = - \log \frac{\exp(\mathrm{sim}(V_i, T_i)/\tau)}{\sum_{j=0}^N \exp(\mathrm{sim}(V_i, T_j)/\tau)}.
    \label{eqn:contra}
\end{equation}

Note this loss does not inject token or word-level supervision, but mainly focuses on learning a joint representation space for images and text, where those that are paired together in the training data would ideally be close to each other in the latent embedding space. 

Surprisingly, we find CLIP trained with paired images and queries leads to a fine-grained understanding of phrase representations, with a simple and effective prompting method which we introduce below.

\subsection{Domain-Aware Prompting}
After pre-training with large-scale image and text data, CLIP can be readily leveraged for image classification via prompting.
Based on prompt engineering in the original paper of CLIP~\citep{radford2021learning}, ``A photo of a [label]'' is a good default template which helps specify that the referring text is about the content of an image. This template is able to improve the zero-shot performance of CLIP for image classification.
We first follow the same template design to prompt CLIP, which we empirically show can also greatly improve the performance for phrase understanding tasks.

However, simply using this template could lead to sub-optimal representations, as the semantics of phrases vary vastly by domain. 
A recent work~\citep{zhou2021learning} has found that adding a domain keyword for a dataset can improve the image classification performance of CLIP.
These domain keywords are usually hand-crafted; to automate this process and build robust phrase representations across different domains, we probe the knowledge of a language model to identify domains for each phrase and design an automatic approach to generating instance-level domain-aware keywords. 

Formally, given a phrase $p_i$, we use ``$p_i$ is a [mask]'' as the template and ask a language model~(e.g., BERT) to fill in the mask token.
We then use the top-K predictions \{$d_i^1$, $d_i^2$, ..., $d_i^K$\} from the language model to construct a prompt ``A photo of $p_i$. A $d_i^1$, ..., $d_i^K$'' for CLIP.
For example, as shown in~\cref{fig:prompt}, given the phrase ``United States'', the language model will generate keywords such as country, republic, and nation. Those keywords then form a domain-aware prompt ``A photo of United States. A country, republic, nation'' that is used as input for CLIP text encoder.

\begin{table*}[t]
\centering  
\small
\begin{tabular}{rccccccccc}
\toprule
\multicolumn{1}{c|}{Datasets}        & \multicolumn{1}{c|}{-} & \multicolumn{2}{c|}{\textbf{CoNLL2003}}   & \multicolumn{2}{c|}{\textbf{BC5CDR}}      & \multicolumn{2}{c|}{\textbf{W-NUT2017}} 
& \multicolumn{2}{c}{\textbf{Avg.}}\\ 
\midrule
\multicolumn{1}{c|}{Metrics}   & \multicolumn{1}{c|}{\# parameters}     & ACC            & \multicolumn{1}{c|}{NMI} & ACC            & \multicolumn{1}{c|}{NMI} & ACC            & \multicolumn{1}{c|}{NMI} &  ACC               & NMI         \\ 
\midrule
\multicolumn{10}{c}{\textit{Vanilla Phrases as Inputs}} \\ 
\midrule
\multicolumn{1}{r|}{Glove}    & -  & 0.528          & 0.166                    & 0.587          & 0.026                 & 0.368             & 0.188   & 0.494             & 0.127          \\
\multicolumn{1}{r|}{BERT} & 110M      & 0.421          & 0.021                    & 0.857          & 0.489               & 0.270             & 0.034       & 0.516            & 0.181      \\
\multicolumn{1}{r|}{LUKE}  & 274M          & 0.478         & 0.093                    & 0.545          & 0.006              & 0.275             & 0.026    & 0.432            & 0.042        \\
\multicolumn{1}{r|}{DensePhrase} & 110M   & 0.388          & 0.037                    & {\ul 0.921}          & 0.616      & 0.268             & 0.061 &   0.526  &  0.238  \\
\multicolumn{1}{r|}{Phrase-BERT} & 110M    & 0.643          & 0.297                    & 0.918          & {\ul 0.617}       & 0.452             & 0.241    & {\ul 0.671}             & {\ul 0.385}        \\
\multicolumn{1}{r|}{UCTopic} & 274M  & 0.485   & 0.081      &  0.776   & 0.320       & 0.322   & 0.074  & 0.528  & 0.158    \\ 
\multicolumn{1}{r|}{CLIP} & 38M & {\ul 0.664} & {\ul 0.431} & 0.560 & 0.008  & {\ul 0.555} & {\ul 0.279} & 0.593 & 0.239 \\
\midrule
\multicolumn{10}{c}{\textit{Prompted Contextual Phrases as Inputs}} \\ 
\midrule
\multicolumn{1}{r|}{BERT} & 110M & 0.725 & 0.387 & 0.852 & 0.456  & 0.386 & 0.206 & 0.654 & 0.350\\
\multicolumn{1}{r|}{LUKE}  & 274M          & 0.739          & 0.406                    & 0.927          & 0.646              & 0.437             & 0.215    & 0.701            & 0.422         \\
\multicolumn{1}{r|}{DensePhrase} & 110M   & 0.637          & 0.299                    & 0.922          & 0.637     & 0.489             & 0.202      & 0.683             & 0.379       \\
\multicolumn{1}{r|}{Phrase-BERT} & 110M    & 0.738         & 0.397                    & 0.918          & 0.624       & 0.454             & 0.215    & 0.703             & 0.355        \\
\multicolumn{1}{r|}{UCTopic} & 274M    & 0.731         & 0.395                    & 0.925          & 0.645       & 0.431             & 0.225    & 0.696             & 0.422        \\
\multicolumn{1}{r|}{CLIP}&38M & \textbf{0.776} & \textbf{0.465} & \textbf{0.930} & \textbf{0.644} & \textbf{0.594} & \textbf{0.303} & \textbf{0.767} & \textbf{0.470}\\
\bottomrule
\end{tabular}

\caption{Performance of entity clustering on three datasets. We compare all models with two differet settings: (1) \textit{Vanilla Phrases as Inputs}: Feed the model with phrases $p_i$; (2) \textit{Prompted Contextual Phrases as Inputs}: Our method of instance-level domain-aware prompting that generate individual keywords \{$d_i^1$, ..., $d_i^K$\} for each phrase $p_i$.}
\label{tab:entity-cluster}
\end{table*}

\subsection{Phrase Understanding}
After designing a prompting method, we can directly use the text encoder of CLIP for phrase understanding, by feeding it with prompted phrases and using the output encodings as phrase representations. We test our method on two tasks, entity clustering and entity set expansion. For entity clustering, we use k-means to cluster the representations. For entity set expansion, given an initial seed of $k$ entities, we average their embeddings and find the nearest entities in the embedding space.

\section{Experiments}
\subsection{Experimental Setup}
\paragraph{Datasets} 
For entity clustering, we conduct experiments on three datasets with labeled entities and semantic categories that cover general, news and biomedical domains as in~\cite{Li2022UCTopicUC}: (1) CoNLL2003~\cite{Sang2003IntroductionTT}, consists of 20,744 sentences from news articles.
(2) BC5CDR~\cite{Li2016BioCreativeVC} is the BioCreative V CDR task corpus from PubMed articles.
(3) W-NUT 2017~\cite{derczynski-etal-2017-results} includes 5,690 sentences and six kinds of entities.
For entity set expansion, we use the WIKI dataset~\cite{shen2017setexpan}
with 8 semantic classes. Each semantic class has 5 seed sets and each seed set contains 3 entities. More details regarding datasets are in~\Cref{app:dataset}.

\paragraph{Evaluation Metrics}
For entity clustering, we follow previous clustering works~\cite{Xu2017SelfTaughtCN} and adopt Accuracy (ACC) and Normalized Mutual Information (NMI) to evaluate different approaches. For entity set expansion, we evaluate the results using Mean Average Precision (MAP) at different top $K$ positions (MAP@K) as below:
\begin{equation}
    \mathrm{MAP}@K = \frac{1}{|Q|}\sum_{q\in Q}\mathrm{AP}_{K}(L_q, S_q),
\end{equation}
where $Q$ is the set of seed queries; $L_q$ and $S_q$ are ranked lists of entities from the model and a ground-truth set respectively; $\mathrm{AP}_{K}(L_q, S_q)$ denotes the average precision at position $K$ for each query $q$.


\paragraph{Baselines}
To show the effectiveness of CLIP for phrase understanding, we compare our method to general language understanding models as well as those specifically trained for phrase understanding:
\begin{itemize}[leftmargin=*,nosep]
    \item \textbf{Glove}~\cite{Pennington2014GloVeGV}. Pre-trained word embeddings. We use averaging word embeddings (dimension $300$) as the representations of phrases.
    \item \textbf{BERT}~\cite{Devlin2019BERTPO}. Phrase representations are produced by averaging token representations (BERT-Avg.) or substituting phrases with the \texttt{[MASK]} token, and using \texttt{[MASK]} representations as phrase embeddings (BERT-Mask).
    \item \textbf{LUKE}~\cite{Yamada2020LUKEDC}. A language model  pre-trained by distant supervision which outputs span representations in sentences. 
    \item \textbf{DensePhrase}~\cite{Lee2021LearningDR}. Learns phrase representations from the supervision of reading comprehension tasks for question answering problems.
    \item \textbf{Phrase-BERT}~\cite{Wang2021PhraseBERTIP}. Context-agnostic phrase representations from contrastive pre-training.
    \item \textbf{UCTopic}~\cite{Li2022UCTopicUC}. Contextual phrase representations from unsupervised contrastive learning that achieves state-of-the-art.
    \item \textbf{CLIP}~\cite{radford2021learning}. We use phrase mentions as inputs of CLIP text encoder to obtain phrase representations.
\end{itemize}

\subsection{Performance Comparison}
\paragraph{Entity Clustering}
First, as shown in~\cref{tab:entity-cluster}, when feeding all models with vanilla phrases without contexts, CLIP already shows competitive or better overall performance compared with most of the baselines. 
Specifically, CLIP performs well on general domain data (CoNLL2003, W-NUT2017), while shows poor results on biomedical domain (BC5CDR) which might appear less in its pre-training data. This can be addressed by adding our domain-aware prompts.
Under the same setting of our prompting method, though all models can benefit from the contextual prompts for phrase understanding, CLIP can significantly outperform BERT by a large margin and achieves even stronger results compared with specialized language models such as Phrase-BERT or UCTopic. Our prompting method also improve the generalizability for phrase understanding across domains. Notably, CLIP also has a much smaller model size compared with large pre-trained language models listed in~\cref{tab:entity-cluster}.

\begin{table}
\small
\centering
\setlength{\tabcolsep}{2pt}
\begin{tabular}{r |cccc}
\toprule
  \multicolumn{1}{c|}{Dataset}  & \multicolumn{3}{c}{\textbf{WIKI}}\\
\cmidrule(lr){1-4}
 \multicolumn{1}{c|}{Metrics} & MAP@10 & MAP@30  & MAP@50 \\
\toprule
\multicolumn{4}{c}{\textit{Vanilla Phrases as Inputs}} \\ 
\midrule
Glove & 0.223 & 0.187 & 0.097\\ 
BERT & 0.608 & 0.536 & 0.418 \\
CLIP & 0.596 & 0.537 & 0.423 \\
\midrule
\multicolumn{4}{c}{\textit{Prompted Contextual Entities as Inputs}} \\ 
\midrule
BERT &  0.610  &  0.545  & 0.456 \\
LUKE & 0.592 & 0.535 & 0.410 \\
DensePhrase  & 0.542  &  0.473  &  0.388 \\  
Phrase-BERT  &   0.636   &   0.579  & 0.492 \\
UCTopic & 0.628 & 0.560 & 0.447 \\ 
CLIP & \textbf{0.739} & \textbf{0.686} & \textbf{0.575}\\
\bottomrule
\end{tabular}
\caption{ Performance comparison on the entity set expansion task.
}
\label{tab:entity-set}
\end{table}

\paragraph{Entity Set Expansion} Results are presented in~\cref{tab:entity-set}. Empirically, we find that BERT performs better for this task than phrase specific language models, potentially due to its exposure to English Wikipedia during pretraining. Nevertheless, our prompted CLIP consistently outperforms the best performing language models for this task.
Notably, CLIP with solely phrases as inputs (MAP@10=0.596) can already outperform some language models (e.g. LUKE, Densephrase) with contextual prompts.

\subsection{Analysis}
\begin{figure}[t]
  \centering
  \includegraphics[width=\linewidth]{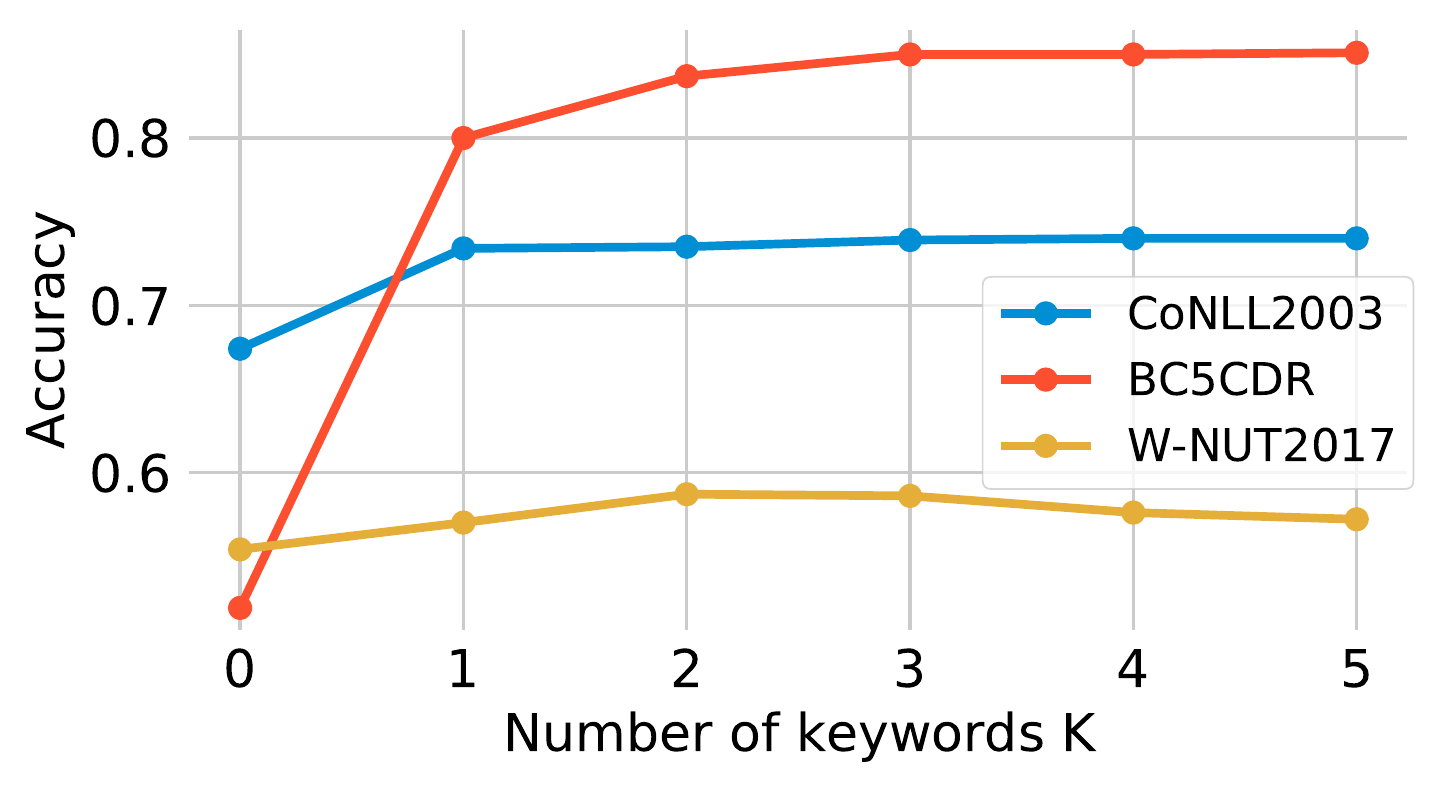}
  \caption{The effect of different number of keywords.}
  \label{fig:acc}
\end{figure}

\paragraph{Sensitivity Tests}
We conduct sensitivity tests on the number of keywords used for prompting. As shown in~\cref{fig:acc}, adding domain-aware keywords ($0\rightarrow K, K\geq1$) can substantially improve the performance, especially for biomedical domain. One possible reason is that biomedical entities are out-of-domain in the training data of CLIP, but our domain-aware prompts can help CLIP understand these entities better.
Overall, the results are robust to the number of keywords added.

\paragraph{Why prompting works?} 
To better understand the mechanism behind our method, we compute visual grounding ratios following~\citep{tan2020vokenization} (see~\Cref{app:visual}) in~\cref{tab:grounding-ratio}. Generated keywords are more visually grounded than original phrases, hence could improve performance as CLIP is trained with visual supervision. Similarly, \citet{hsu2022xdbert} also find CLIP is better at tasks that are more visually grounded.

\begin{table}
\small
\centering
\setlength{\tabcolsep}{2pt}
\begin{tabular}{r | ccc}
\toprule
  \multicolumn{1}{c|}{Dataset}  & \multicolumn{1}{c|}{\textbf{CoNLL2003}}   & \multicolumn{1}{c|}{\textbf{BC5CDR}}      & \multicolumn{1}{c}{\textbf{W-NUT2017}} \\
\toprule
Phrases & 0.59\% & 0.18\% & 1.75\% \\
\midrule
Keywords & 13.96\% & 7.25\% & 21.10\% \\
\bottomrule
\end{tabular}
\caption{ Estimated visual grounding ratios.
}
\label{tab:grounding-ratio}
\end{table}

\section{Related Work}
\paragraph{Pre-trained Models and Prompting}
Pre-trained language models~\citep{Devlin2019BERTPO, radford2019language, brown2020language} have been the dominant methods for natural language processing.
A recent trend to efficiently leverage these powerful language models is to design textual prompts~\citep{houlsby2019parameter, shin2020autoprompt, gao2020making, scao2021many}. 
Recent work also explores prompting methods for CLIP~\citep{zhou2021learning,song2022clip}, specifically on image classification and multimodal tasks. 
Our work differs in that we focus on studying the text encoder of CLIP alone for text understanding tasks.


\paragraph{Phrase Understanding}
Phrase understanding is an important task for NLP. Early works~\cite{Yu2015LearningCM, Zhou2017LearningPE} combine word embeddings to get phrase representations. Phrase-BERT~\cite{Wang2021PhraseBERTIP} composed token embeddings from BERT and conducted contrastive pretraining on datasets constructed by a GPT-2-based diverse paraphrasing model~\cite{Krishna2020ReformulatingUS}. \citet{Lee2021LearningDR} used the supervision from reading comprehension tasks to learn dense phrase representations for QA tasks. 
In this work, we bring a new perspective to study visually supervised text encoder for phrase understanding.

\section{Conclusion}
In this paper, we show
that the text encoder of CLIP, which is trained with a single image-text contrastive loss, can provide good phrase representations compared with strong language models that are trained with masked language modeling. 
We can further improve the representations with a simple and effective prompting method.
In the future, it would be interesting to explore CLIP for more text understanding tasks, e.g., semantic text similarity.

\bibliography{anthology,emnlp2022}
\bibliographystyle{acl_natbib}

\clearpage
\appendix

\section{Appendix}
\label{sec:appendix}
\subsection{Implementation Details}
All the pre-trained models used in the experiments are publicly available.  We use their official release to extract phrase representations. We use CLIP version with ViT-B/32 as the visual encoder~\footnote{\url{https://github.com/openai/CLIP}}, with 37.8M parameters. For BERT model, we use BERT-base-uncased~\footnote{\url{https://huggingface.co/models}} with 110M parameters. We set the number of keywords $K$ to 3 as it achieves the best accuracy on average (shown in~\cref{fig:acc}). We use BERT-large-cased to generate keywords for each phrase. We use PyTorch and HuggingFace to load the models, all codes in our experiments are implemented on a NVIDIA TITAN RTX GPU server. All representations are extracted within 5 minutes per dataset. All evaluation methods are based on Scikit Learn metrics package~\footnote{\url{https://scikit-learn.org/stable/modules/model_evaluation.html}}.

\subsection{Dataset Details}
\label{app:dataset}
(1) CoNLL2003~\cite{Sang2003IntroductionTT}, consists of 20,744 sentences extracted from Reuters news articles. The entity categories are Person, Location and Organization. Because Misc category does not represent a single semantic category, we do not include Misc entities in our experiments.
(2) BC5CDR~\cite{Li2016BioCreativeVC} is the BioCreative V CDR task corpus. It has 18,307 sentences from PubMed articles, with 15,953 chemical and 13,318 disease entities. 
(3) W-NUT 2017~\cite{derczynski-etal-2017-results} contains unusual entities in the context of emerging discussions and we include 5,690 sentences and six kinds of entities: corporation, creative work, group, location, person, product in our experiments.
(4) WIKI~\cite{shen2017setexpan} includes 33K entities from 8 semantics classes, i.e., china provinces; companies; countries; diseases; parties; sports leagues; tv channels; US states.

\subsection{Visual Grounding Ratio}
\label{app:visual}
We follow~\citep{tan2020vokenization} to estimate visual grounding ratios. Visually grounded words are those that are naturally related to specific visual concepts (e.g. cat, river, ball), Since a precise list of such words are hard to define, we consider a word with more than 100 occurances in the MS-COCO dataset as visually-grounded. The reason to use MS COCO is that it is an image captioning dataset where words in the text are considered to be mostly visually grounded. By doing so, one can roughly estimate the visual grounding ratio for text datasets, and the results are compliant with epirical findings~\citep{tan2020vokenization, hsu2022xdbert}.
\end{document}